\documentclass{style/nseJournal}

\usepackage{booktabs}
\usepackage{subcaption}
\usepackage{amsmath,amssymb}
\usepackage{graphicx}
\usepackage{url}
\usepackage{gensymb}
\usepackage{xcolor}
\usepackage{soul}
\usepackage[percent]{overpic}
\newcommand{\vr}[1]{{{#1}}}

\begin{document}

\title{An Automated and Reproducible Workflow for Crack Identification and Damage Assessment of Fusion Materials}

\addAuthor{\correspondingAuthor{Rinkle Juneja$^\dagger$}}{a}
\correspondingEmail{junejar@ornl.gov}
\addAuthor{Viktor Reshniak$^\dagger$}{a}
\addAuthor{Richard K. Archibald}{a}
\addAuthor{John W. Duggan}{a}
\addAuthor{Gregory R. Watson}{a}
\addAuthor{Cory D. Hauck}{a}
\addAuthor{Gary M. Staebler}{a}
\addAffiliation{a}{Oak Ridge National Laboratory, Oak Ridge, TN 37831}

\addKeyword{Fusion materials}
\addKeyword{Galaxy workflows}
\addKeyword{Image processing}
\addKeyword{Crack identification}
\addKeyword{Scanning electron microscopy}
%\addKeyword{Data provenance}

\titlePage
\vspace{-1em}
\noindent{\small $^\dagger$ Contributed equally}

\begingroup
\renewcommand{\thefootnote}{}
\footnotetext{\scriptsize
This manuscript has been authored by UT-Battelle, LLC, under contract DE-AC05-00OR22725 with the U.S. Department of Energy (DOE).
The U.S. Government retains and the publisher, by accepting the article for publication, acknowledges that the U.S. Government retains a nonexclusive, paid-up, irrevocable, worldwide license to publish or reproduce the published form of this manuscript, or allow others to do so, for U.S. Government purposes.
DOE will provide public access to these results of federally sponsored research in accordance with the DOE Public Access Plan (\url{https://www.energy.gov/doe-public-access-plan}).
}
\addtocounter{footnote}{-1}
\endgroup

\begin{abstract}
Post-exposure microscopy is central to qualification of fusion materials. However, manual analysis does not scale to the volume, heterogeneity, and multiresolution character of modern fusion-materials campaigns. To address this challenge, we present a reproducible workflow, implemented in the Galaxy scientific workflow environment, for automated crack identification and quantitative damage assessment from scanning electron microscopy images. 
% The workflow converts microscopy data and experimental metadata into provenance-tracked crack masks, skeletonized crack networks, quality-control visualizations, and scalar damage descriptors. 
\vr{The workflow processes SEM images and experimental metadata to identify cracks, quantify damage, and retain the intermediate products and processing history needed for reproducibility. Outputs include crack masks, skeletonized crack networks, quality-control visualizations, and scalar damage descriptors.}
The method is designed to operate without image-specific parameter tuning across tungsten grades, microstructures, magnifications, and damage states. 
We demonstrate the workflow on a sparse electron-beam thermal-shock dataset containing 418 images from 114 experiments spanning five tungsten grades and three microstructural states. We define a crack-density descriptor, which provides standardized inputs for downstream machine-learning prediction and physics-based crack simulation. These predictive components are exposed in the same Galaxy environment and are intentionally treated here as extensible workflow modules. The principal contribution is therefore an end-to-end, shareable, and computationally portable workflow that links experimental characterization, automated image analysis, preliminary damage prediction, and simulation-guided data acquisition for fusion-materials research.
\end{abstract}

\section{Introduction}
Plasma-facing and structural materials in fusion systems must operate under combinations of high heat flux, steep thermal gradients, cyclic loading, plasma exposure, and particle irradiation.
These conditions can drive surface roughening, recrystallization, plastic deformation, embrittlement, crack initiation, and crack-network growth \cite{brezinsek2017,linke2011,wirtz2016}. 
Reliable material selection and component qualification therefore require not only high-fidelity experiments and simulations, but also repeatable and automated methods for extracting quantitative damage information from post-exposure characterization \cite{JUNEJA2026156473}.

The Material Plasma Exposure eXperiment (MPEX) facility, currently under construction at Oak Ridge National Laboratory, will generate diverse pre- and post-exposure measurements of plasma-material interaction damage, including cracking, sputtering, erosion, and redeposition \cite{rapp2020,rapp2023finaldesign,Lau_2023}.
Targets will be transported under vacuum to the MPEX Surface Analysis Station, which will include focused-ion-beam milling, high-resolution scanning electron microscopy (SEM), surface-mapping \vr{X-ray} spectroscopy, and complementary diagnostics \cite{BiewerRapp2020SAS,rapp2020}. 
This combination will produce a large and heterogeneous characterization stream that must be converted into consistent material-response descriptors before it can support model validation, candidate-material comparison, and data-driven experiment planning.

%\begin{figure}[tb]
%    \centering
%    \includegraphics[scale=0.85]{figures/MPEX.pdf}
%    \caption{Drawing of the Material Plasma Exposure eXperiment (MPEX) showing the Helicon plasma source, Electron Cyclotron Heating (ECH), Ion Cyclotron Heating (ICH) and Plasma Material Interaction (PMI) regions }
%    \label{fig:MPEX}
%\end{figure}

SEM is particularly valuable for probing the crack damage because it resolves crack networks and surface modification across a wide range of length scales. 
Yet SEM datasets are difficult to analyze consistently.
Crack contrast depends on detector response and local topography; cracks may appear dark or bright; polishing marks, grain boundaries, pores, redeposited material, and roughened surfaces can produce crack-like intensity features; and a fixed number of pixels can represent substantially different physical dimensions across images.
Manual inspection remains essential for expert interpretation, but it is time-consuming, subjective, and difficult to reproduce across analysts or campaigns \cite{Zhao2021SEMCrack,MohanPoobal2018}.

The challenge is broader than segmentation. 
A useful material-assessment capability must connect raw images with experimental metadata, record each processing operation applied to the data, including intensity rescaling, image enhancement, thresholding, morphological refinement, and descriptor calculation, preserve intermediate outputs such as enhanced images, histograms, selected thresholds, binary masks, overlays, and skeletons, support re-execution on new data, and provide standardized outputs to statistical-learning and physics-simulation tools. 
Stand-alone scripts can perform individual operations, but they often depend on local software environments, undocumented preprocessing choices, and manual file transfer. 
Such practices hinder traceability and make it difficult to compare results across facilities and teams.

This paper presents a workflow-centered solution.
We integrate automated crack identification into Galaxy, a web-based scientific workflow platform that manages tools, data movement, execution histories, parameter settings, and computational resources \cite{GalaxyCommunity2024,Goecks2010}. 
The workflow accepts SEM images and associated metadata, performs automated crack extraction, computes quantitative descriptors, generates quality-control products, and makes the results available to downstream machine-learning and fracture-simulation modules. 
The same workflow can be executed interactively, shared with collaborators, invoked through an application programming interface, or dispatched to local and high-performance computing resources \cite{GalaxyCommunity2024,Watson2024GalaxyNeutrons}.

The emphasis is intentionally placed on workflow definition, reproducibility, and the automatic crack-identification component. 
Preliminary machine-learning and simulation capabilities are included to establish the interfaces of the broader material-assessment workflow, while their detailed formulation and validation are reserved for follow-on papers.

\vr{The rest of the manuscript is organized as follows. Section \ref{sec:data} describes the experimental dataset and workflow requirements. Section \ref{sec:galaxy} introduces the Galaxy architecture. Sections \ref{sec:cracks} and \ref{sec:workflow} describe crack identification and quality control. Section \ref{sec:modeling} describes interfaces to downstream modeling. Section \ref{sec:discussion} concludes with a discussion on limitations and the path toward closed-loop assessment.}

\section{Experimental Dataset and Workflow Design}
\label{sec:data}
\subsection{Electron-beam thermal-shock dataset}
The demonstration dataset consists of post-exposure SEM images from electron-beam thermal-shock experiments performed at the JUDITH~1 facility across a range of base temperatures and heat fluxes \cite{wirtz2013,linke2011,wirtz2016}.
Electron-beam exposure is used here as a thermal-only surrogate condition that provides material-response data before MPEX operation. 
Five tungsten grades are included: nominally pure tungsten (pure W, 99.97\%), ultra-high-purity tungsten (W-UHP, 99.9999\%), potassium-doped tungsten (WVMW), tungsten with 1\% tantalum (WTa1), and tungsten with 5\% tantalum (WTa5). 
For each composition, forged samples were cut in longitudinal and transverse orientations, and additional samples were recrystallized, producing 15 composition-microstructure combinations~\cite{wirtz2013}.

\begin{figure}[tbp]
    \centering
    \begin{subfigure}[t]{0.45\textwidth}
        \vspace{0pt}
        \centering
        \footnotesize
        \setlength{\tabcolsep}{6pt}
        \renewcommand{\arraystretch}{1.45}

        \begin{tabular}{lcccc}
        \toprule
        & & \multicolumn{2}{c}{\# images} \\
        \cmidrule(lr){3-4}
        Material & \# experiments & total & L / R / T \\
        \midrule
        All            & 114 & 418 & 269 / 71 / 78 \\
        M192 -- W-UHP  & 32  & 100 & 77 / 11 / 12 \\
        M193 -- WVMW   & 23  & 76  & 52 / 11 / 13 \\
        M194 -- WTa1   & 17  & 67  & 39 / 12 / 16 \\
        M195 -- WTa5   & 19  & 83  & 33 / 24 / 26 \\
        M196 -- pure W & 23  & 92  & 68 / 13 / 11 \\
        \bottomrule
        \end{tabular}
        \caption{Dataset summary by material.}
        \label{fig:mat:stat}
    \end{subfigure}
    \hfill
    \begin{subfigure}[t]{0.44\textwidth}
        \vspace{0pt}
        \centering
        \includegraphics[width=\textwidth]{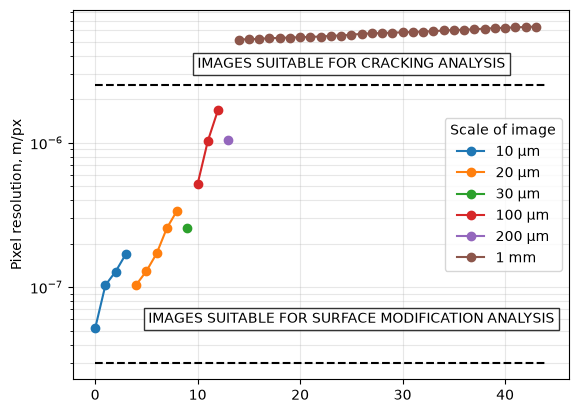}
        \caption{Physical pixel resolution grouped by nominal acquisition scale.} 
        \label{fig:mat:resolutions}
    \end{subfigure}
    \\[1.5em]
    \begin{subfigure}[t]{\textwidth}
        \vspace{0pt}
        \centering
        \includegraphics[width=0.32\textwidth]{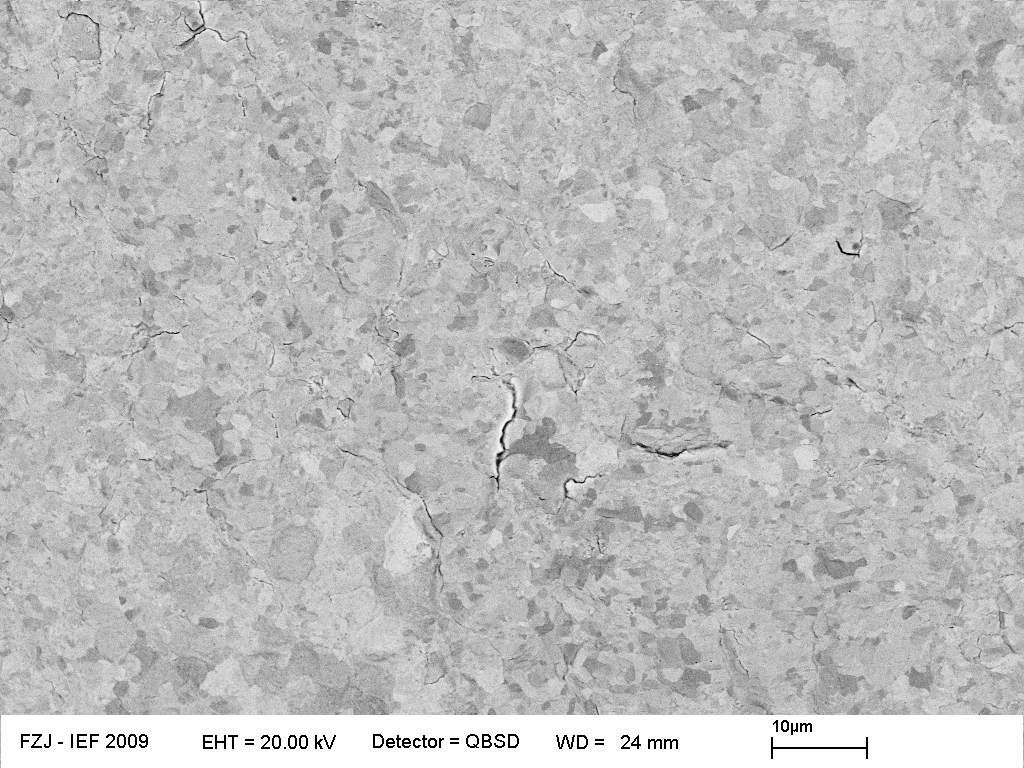}
        \hfill
        \includegraphics[width=0.32\textwidth]{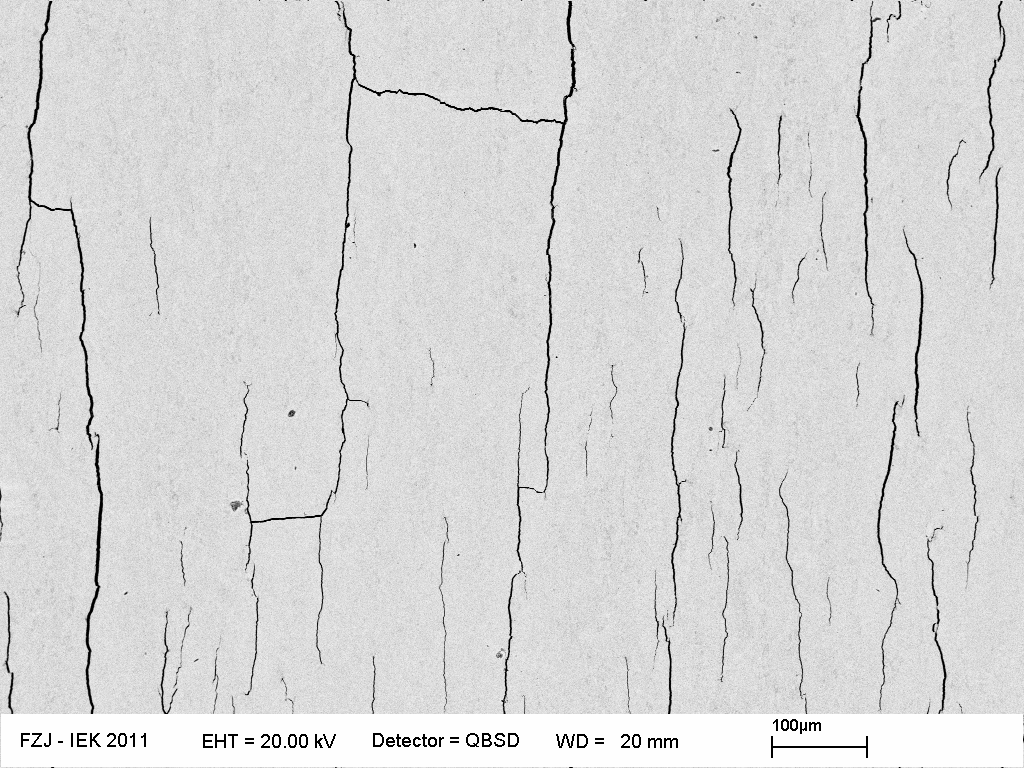}
        \hfill
        \includegraphics[width=0.32\textwidth]{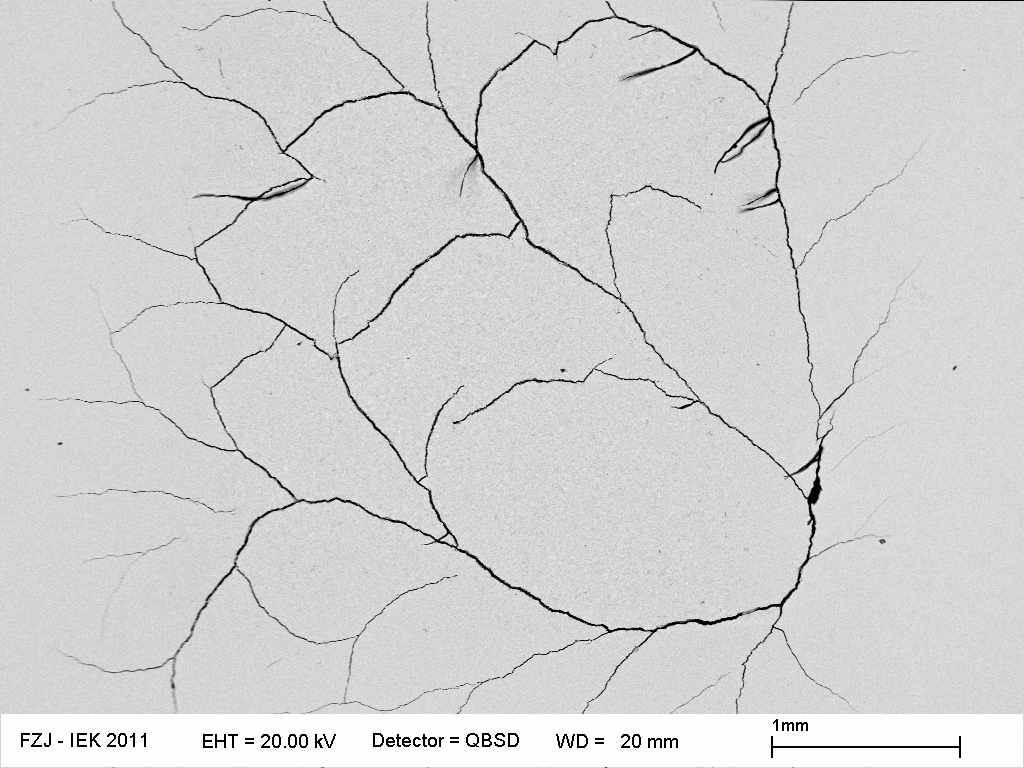}
        \caption{Representative surface scans acquired with nominal scale labels of $10\mu m$, $100\mu m$ and $1 mm$}
        \label{fig:mat:img_examples}
    \end{subfigure}
    \\[1.5em]
    \begin{subfigure}[t]{\textwidth}
        \vspace{0pt}
        \centering
        \includegraphics[width=\textwidth]{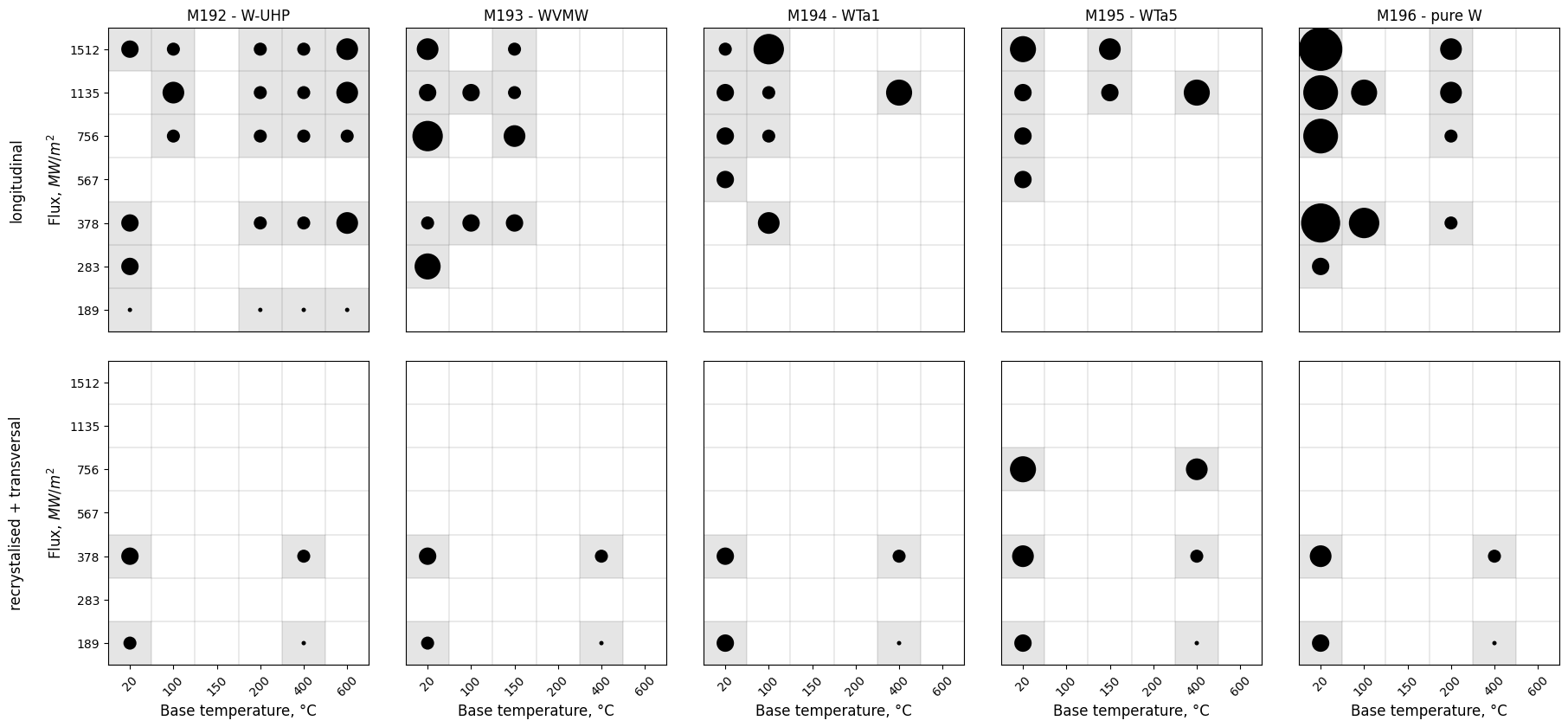}
        \caption{Test matrix of the dataset with marker size showing the relative number of collected images.}
        \label{fig:mat:test_mat}
    \end{subfigure}
    \caption{Overview of the collected surface-scan dataset. 
    (a)~Number of experiments and images per material, with image counts split into longitudinal (L), recrystallized (R), and transverse (T) cuts. 
    (b)~Pixel resolution distribution across image scales. 
    (c)~Example scans at fine and coarse scales.
    (d)~Dataset coverage across materials and exposure conditions.
    }
    \label{fig:mat:overview}
\end{figure}

Across these material states, 114 experiments generated 418 SEM images. 
\vr{The number of images associated with each experiment depended on the number of surface locations and acquisition scales imaged for that specimen.}
Expert inspection assigned qualitative outcomes such as cracking, surface modification, or no apparent damage. 
\vr{Figure~\ref{fig:mat:stat}} summarizes the image distribution. 
The test matrix is sparse and unevenly sampled, with fewer observations for transverse and recrystallized materials than for longitudinal material. 
% This imbalance is partly purposeful: materials that failed at lower heat flux or base temperature were not always tested extensively at more severe conditions. 
\vr{The smaller number of transverse and recrystallized observations partly reflects the progression of the original test campaign: for several of these material states, failure occurred at lower heat flux or base temperature, and testing was therefore not continued extensively at more severe conditions.}
The resulting sparsity motivates a workflow in which robust characterization is treated as a mature component and prediction is treated more cautiously.

% \begin{table}[tb]
% \centering
% \caption{Summary of the electron-beam post-exposure SEM dataset. Image counts are split into longitudinal (L), recrystallized (R), and transverse (T) microstructures.}
% \label{tab:dataset}
% \begin{tabular}{lrrc}
% \toprule
% Material & Experiments & Images & L/R/T \\
% \midrule
% All & 114 & 418 & 269/71/78 \\
% W-UHP & 32 & 100 & 77/11/12 \\
% WVMW & 23 & 76 & 52/11/13 \\
% WTa1 & 17 & 67 & 39/12/16 \\
% WTa5 & 19 & 83 & 33/24/26 \\
% Pure W & 23 & 92 & 68/13/11 \\
% \bottomrule
% \end{tabular}
% \end{table}

% Images span several magnifications and physical fields of view, from fine-scale microstructural observations to millimeter-scale surface surveys. 
% The workflow therefore treats pixel resolution and image scale as first-class metadata. 
% Images that are too coarsely resolved to support crack-morphology analysis can be retained for surface-modification studies, but they are flagged or excluded from crack-specific operations.
Images were acquired using several nominal scale settings recorded in the image metadata and displayed as labeled reference line segments, such as $10\mu\mathrm{m}$, $100\mu\mathrm{m}$, or $1\mathrm{mm}$ as shown in Figure~\ref{fig:mat:img_examples}. 
Because these scale labels do not directly provide a consistent measure of spatial sampling, the workflow converts each image scale to a physical pixel resolution, defined as the physical length represented by one pixel. 
This derived resolution is used to compare images acquired under different scale settings and to determine whether an image is sufficiently resolved for crack morphology analysis as shown in Figure~\ref{fig:mat:resolutions}.
Images that are too coarsely resolved can be retained for surface modification studies, but they are flagged or excluded from crack specific operations.

\subsection{Workflow requirements}
The workflow was designed around five requirements relevant to fusion-materials campaigns: automation without image-by-image threshold selection; physical consistency across magnifications; traceability of data, parameters, software versions, and intermediate results; portability across workstation, cloud, and high-performance computing resources; and extensibility so that image-analysis products can be consumed directly by statistical-learning and physics-simulation modules.

\section{Galaxy Workflow Architecture}
\label{sec:galaxy}
\subsection{Galaxy as the orchestration layer}
Galaxy is a web-based data-analysis platform used in data-intensive scientific domains such as bioinformatics, neutron scattering, and others \cite{GalaxyCommunity2024,Watson2024GalaxyNeutrons}. 
One of Galaxy's key capabilities is the integration of tools, data, and computational resources while managing provenance and reproducibility \cite{GalaxyCommunity2024,Goecks2010}. 
Individual tools can be composed into workflows that span multiple computing resources, possibly hosted at different sites. 
These workflows can be invoked through a web browser, custom user interfaces, or scripts, and the resulting outputs can be visualized within the Galaxy user interface or exported to external visualization software.

A Galaxy instance has been established at ORNL to deploy MPEX tools and workflows. 
This instance can use dedicated fusion-related computing hardware, internal cloud resources through the ORNL Open Research Cloud, ORNL leadership computing resources such as Frontier, and external DOE resources such as NERSC Perlmutter from a single workflow interface.
This separation between the user-facing workflow and the execution resources is essential for a heterogeneous material-assessment pipeline: lightweight image processing can run on conventional nodes, whereas large simulations can be dispatched to GPU-enabled high-performance computational systems.

\subsection{Tool deployment and data contracts}
Creating a software tool in Galaxy requires two complementary artifacts. 
First, a build of the software is prepared for the hardware on which it will execute. 
Second, an XML wrapper defines an execution command or script, required input files, user-exposed parameters, output files, dependencies, metadata, and help information \cite{Blankenberg2014ToolShed,GalaxyCommunity2024}. 
Most MPEX tools are packaged in platform-independent containers. 
Containerization adds an extra packaging step, but greatly simplifies dependency management and allows the same implementation to run on any compatible system \cite{Boettiger2015Docker,Kurtzer2017Singularity}. 
On HPC systems that restrict direct container execution, Galaxy uses batch-job submission and a native software build on the target system.

\begin{figure}[tb]
  \centering
  \includegraphics[width=\textwidth]{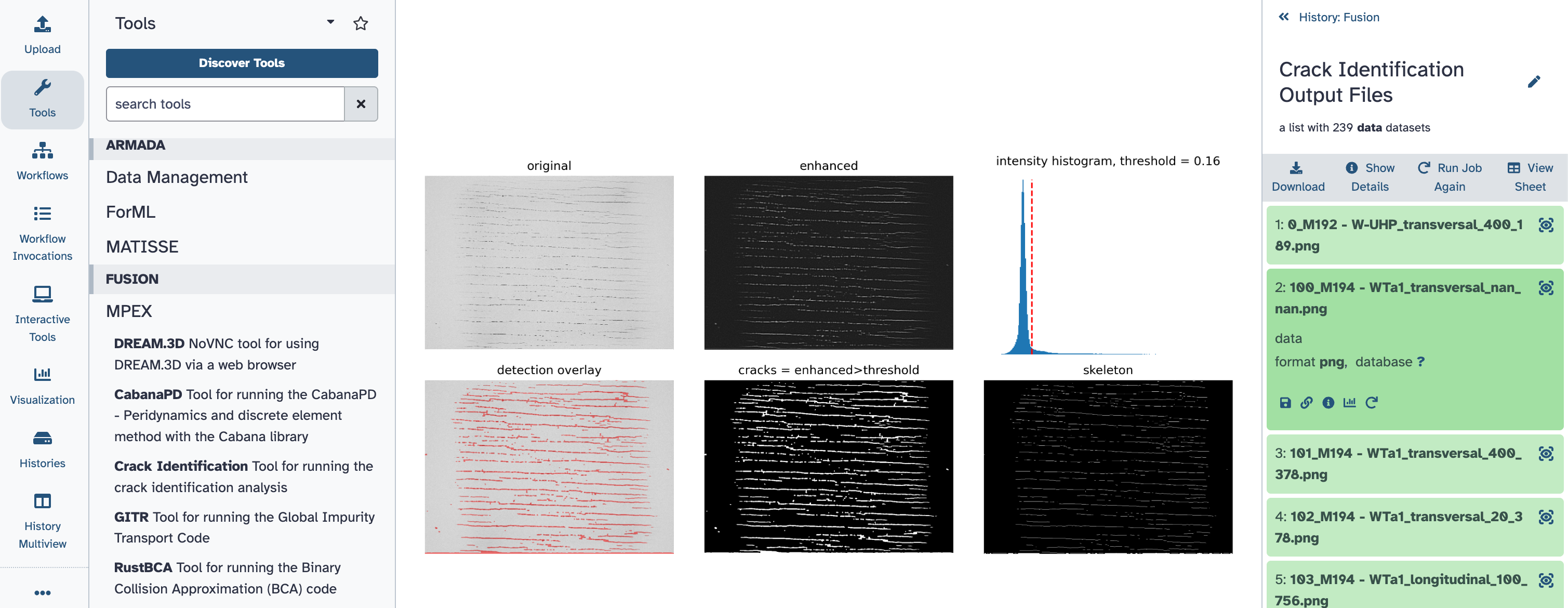}
  \caption{Galaxy display of results from the automated crack identification tool. The interface shows the original microscopy image, the ridge enhanced image used for segmentation, the intensity histogram with the automatically selected threshold, the detected cracks overlaid on the original image, the binary crack mask, and the one pixel wide skeleton of the detected crack network. The Galaxy history shown at the side records the input files, processing steps, parameter settings, and generated outputs.
  % Galaxy visualization of output from the automated crack-identification tool. The interface presents the original image, enhanced image, histogram and threshold, crack overlay, binary mask, and skeleton while retaining the execution history and output files.  
  }
  \label{fig:galaxy:results}
\end{figure}

For our crack-identification tool, the wrapper defines typed inputs for microscopy images, optional image collections when multiple images are processed together, physical pixel size, image scale metadata, and processing parameters. 
Its outputs include enhanced images, binary crack masks, crack-overlay visualizations, scalar descriptors, machine-readable summary tables, and, when requested, one pixel wide skeletons of the detected crack networks.
These standardized outputs allow subsequent tools to use the crack identification results directly without manual reformatting or file conversion.
In particular, the same crack-density table can be used by a regression tool, while binary masks or skeletons can be used by topology-analysis tools.
Figure~\ref{fig:galaxy:results} shows these outputs as presented in the Galaxy interface.

Users can import data through the web upload tool, through scripts that interact with the Galaxy API, or through future automated ingestion from experimental data systems. 
Galaxy maintains a history of all imported files, tool executions, parameters, and outputs. 
Histories can be shared with collaborators and can support reproducible packaging approaches such as RO-Crates~\cite{SoilandReyes2022ROCrate,GalaxyCommunity2024}.

\subsection{Workflow composition}
The principal benefit of Galaxy is its ability to compose individual tools into workflows.
Galaxy automatically transfers outputs from one tool to compatible inputs of subsequent tools and restricts workflow connections to data types that are mutually compatible \cite{GalaxyCommunity2024}. 
A workflow can therefore be configured so that the user supplies only the initial image collection and metadata, even when those files feed several downstream branches.

For the material-assessment use case, the baseline workflow is:
\begin{enumerate}
    \item ingest SEM images and associated experimental metadata;
    \item validate image readability, dimensional metadata, and crack-analysis suitability;
    \item run automated crack identification over the image collection;
    \item generate images for expert review, including enhanced images, crack masks, crack overlays, and skeletonized crack networks, together with quantitative descriptors;
    \item aggregate descriptors by material and exposure condition; and
    \item \vr{optionally pass crack descriptors to statistical-learning tools and compare experimental descriptors with descriptors derived from physics-based simulations.}
    % \item optionally pass standardized outputs to machine-learning prediction or simulation-analysis tools.
\end{enumerate}

Once saved, a workflow can be shared and reused without requiring each user to understand the implementation details or manually move files among tools. 
Advanced users can still inspect or modify the workflow, replace modules, alter parameters, or invoke it programmatically.

\section{Automated Crack Identification}
\label{sec:cracks}
\subsection{Image processing}

\paragraph{Processing challenges.}
Reliable crack identification is difficult because the dataset spans a wide range of resolutions and surface morphologies \cite{Zhao2021SEMCrack,MohanPoobal2018}. Cracks occur at different length scales and with different contrast relative to the background. Depending on local topography and detector response, they may appear either dark or bright. Polishing marks, grain boundaries, and surface modification can create intensity changes comparable to those of actual cracks. A fixed global threshold can therefore miss faint cracks in low-contrast images or over-segment textured regions in high-contrast images.

\begin{figure}[tb]
    \def\w{0.22\textwidth}
    \centering
    \begin{tabular}{cccc}
        \textbf{\small Original} &
        \textbf{\small Sato ridge filter} &
        \textbf{\small Gaussian smoothing} &
        \textbf{\small Unsharp masking} \\[0.5em]
        \includegraphics[width=\w]{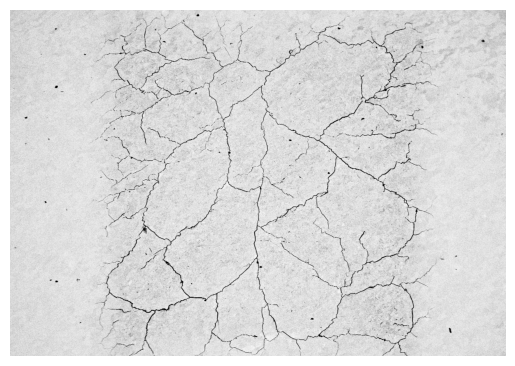} &
        \includegraphics[width=\w]{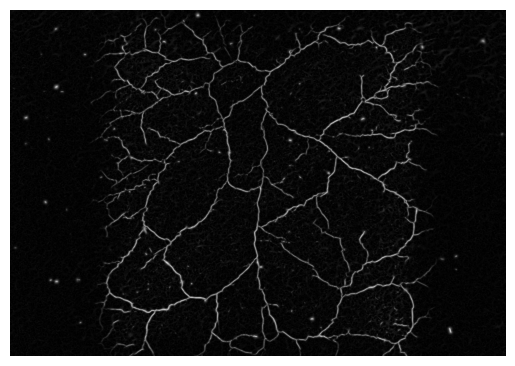} &
        \includegraphics[width=\w]{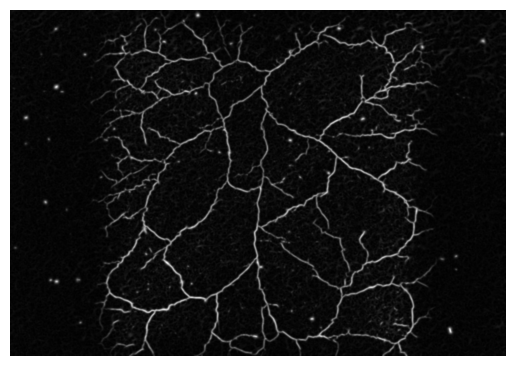} &
        \includegraphics[width=\w]{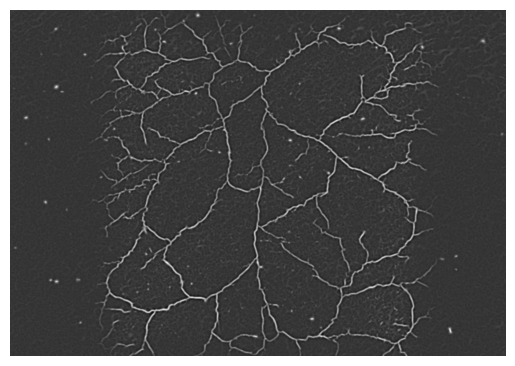} \\[1em]
        \begin{overpic}[width=\w]{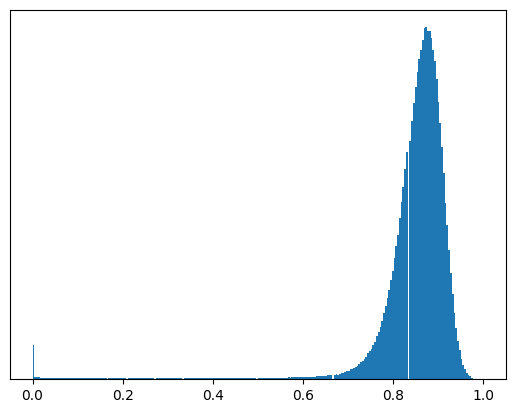}
            \put(60,60){\makebox(0,0)[r]{\scriptsize\shortstack{Intensity gap}}}
            \put(64,55){\vector(1,-1){15}}
        \end{overpic}&
        \begin{overpic}[width=\w]{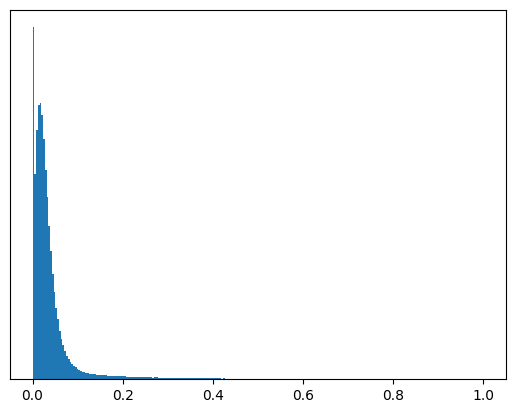}
            \put(25,40){\scriptsize\shortstack{Dominant \\background peak}}
            \put(24,55){\vector(-1,1){15}}
        \end{overpic}&
        \includegraphics[width=\w]{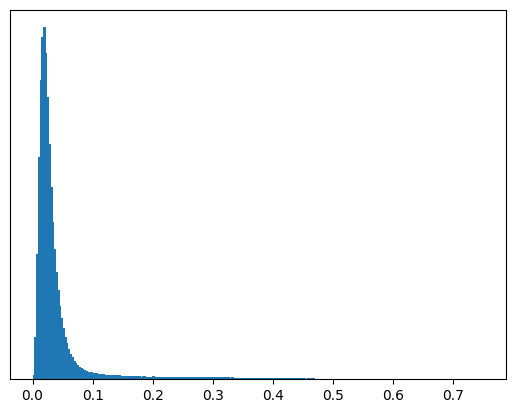} &
        \begin{overpic}[width=\w]{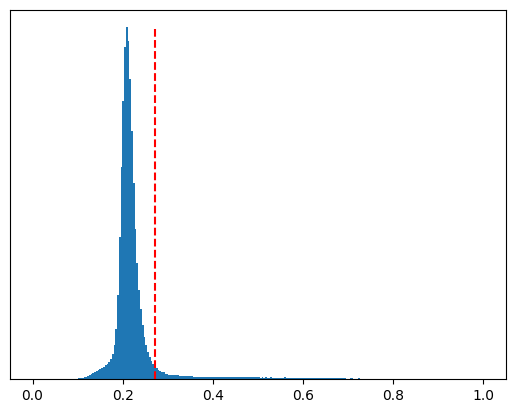}
            \put(50,50){\scriptsize\shortstack{Selected \\threshold}}
            \put(48,48){\vector(-1,-1){15}}
        \end{overpic}
    \end{tabular}
    \caption{
    % Image-processing pipeline for crack detection. Top row: original image and successive enhancement steps. Bottom row: corresponding intensity histograms, showing how the distribution is conditioned for robust thresholding. \cdh{what are the lines.  what does "conditioned for robust thresholding" mean?}
    % \vr{Image processing steps and corresponding intensity histograms. The labeled line in the final histogram marks the automatically selected  threshold used to generate the crack masks shown in Figure~\ref{fig:mat:sem}.}
    Image processing steps and corresponding intensity histograms. In each histogram, the horizontal axis represents normalized pixel intensity and the vertical axis represents the number of pixels. The annotations identify representative image specific intensity artifacts and the automatically selected threshold used to generate the crack masks shown in Figure~\ref{fig:mat:sem}.
    }
    \label{fig:mat:img_preproc}
\end{figure}

% The processing pipeline was designed to enhance elongated structures while suppressing background variability and to perform segmentation without image-specific parameter tuning. Figure~\ref{fig:mat:img_preproc} shows the principal stages and their effect on the intensity distribution.
% \vr{The processing pipeline progressively enhances elongated crack responses and modifies the associated intensity distribution before automatic segmentation. 
% The threshold is calculated from the histogram of the final enhanced image, see Figure~\ref{fig:mat:img_preproc}. 
% Pixels above this threshold are classified as cracks, producing the binary masks and crack overlays shown in Figure~\ref{fig:mat:sem}.}
The processing pipeline progressively enhances elongated crack responses while reducing image specific intensity artifacts that can interfere with automatic segmentation. 
Figure~\ref{fig:mat:img_preproc} shows these processing stages and the corresponding histograms. 
The annotations identify representative artifacts, including gaps in the intensity distribution and a strong concentration of background pixels, that are reduced or redistributed through smoothing and contrast restoration. 
These operations condition the final intensity distribution so that the threshold can separate crack responses from the background more consistently. 
The threshold is selected automatically for each image, and pixels above it are classified as cracks, producing the binary masks and crack overlays shown in Figure~\ref{fig:mat:sem}.

\paragraph{Ridge enhancement.}
% \cdh{I don't see a correlation between this description and the figures in the top row above.}  The first stage applies Sato ridge enhancement, which detects line-like structures from second-order intensity variation through the Hessian matrix. \cdh{what Hessian matrix?}
% Cracks exhibit strong curvature perpendicular to their direction and weaker variation along it. 
% The filter exploits this anisotropy to enhance elongated crack-like structures while suppressing more isotropic background features. 
% A multiscale implementation allows the response to span the range of apparent crack widths represented in the image collection \cite{SATO1998143}.

The second column of Figure~\ref{fig:mat:img_preproc} shows the response after applying the Sato ridge filter to the original image. 
The image is first linearly rescaled to the interval [0,1] using its minimum and maximum intensities. 
The Sato filter then evaluates the Hessian matrix, which contains the second spatial derivatives of a Gaussian smoothed version of the image. 
Its eigenvalues describe how rapidly intensity changes along two perpendicular directions. 
Because an elongated crack produces a strong variation across the crack and a weaker variation along it, the filter enhances crack like ridges while suppressing more isotropic background structures \cite{SATO1998143}. 
The filter is applied at Gaussian scales ($\sigma=1,2,3$) pixels, and the strongest ridge response across these scales is retained. 
The resulting response is rescaled to [0,1]. 
As shown in the second histogram, this operation concentrates much of the background response near low intensity values while assigning larger responses to elongated features.
% \vr{The second column of Figure~\ref{fig:mat:img_preproc} shows the response after applying the Sato ridge filter to the original image. 
% At each selected scale, the filter evaluates the Hessian matrix, which contains the second spatial derivatives of a Gaussian smoothed version of the image. 
% The eigenvalues of this matrix describe how rapidly image intensity changes along two perpendicular directions. 
% An elongated crack produces a strong intensity variation across the crack and a weaker variation along it. 
% The Sato filter uses this directional difference to enhance crack like ridges while suppressing more isotropic background structures \cite{SATO1998143}. 
% The filter is evaluated at Gaussian scales ($\sigma=1,2,$) and ($3$) pixels, and the multiscale response retains the strongest ridge response across these scales. 
% As shown in the second histogram, this operation concentrates much of the background response near low intensity values while assigning larger responses to elongated features.}

\paragraph{Histogram conditioning.}

% The ridge-enhanced image is subjected to Gaussian smoothing. 
% In addition to suppressing high-frequency noise, this step stabilizes the histogram used for automatic thresholding. 
% Ridge enhancement can produce sparse or irregular intensity distributions in which values are concentrated in a small number of bins. 
% Gaussian smoothing redistributes intensities through local averaging, fills histogram gaps, and produces a more continuous distribution.
% An unsharp mask is then applied to restore local contrast attenuated by smoothing \vr{before the threshold is applied.} 
% This enhances crack boundaries by amplifying local intensity differences without fully reintroducing the noise or histogram sparsity present before smoothing \cite{GonzalezWoods2018}.
The third and fourth columns of Figure~\ref{fig:mat:img_preproc} show the conditioning steps applied before thresholding. 
Gaussian smoothing with $\sigma=1$ pixel reduces isolated fluctuations and fills gaps in the ridge response histogram, producing a more continuous intensity distribution. Unsharp masking is then applied with a radius of 2 pixels and a dimensionless gain of 2. 
The gain controls the strength with which local contrast is restored after smoothing. The image is then rescaled to [0,1] before segmentation. 
These operations reduce image specific intensity artifacts while preserving the contrast between enhanced crack responses and the background, thereby producing a more stable histogram for automatic intensity thresholding. 
The same parameter values were used for all images analyzed in this study.

% \paragraph{Implementation details.}
% In the implemented pipeline, each grayscale image is first linearly rescaled to the interval $[0,1]$ based on the minimum and maximum image intensities. 
% Dark crack-like structures are then enhanced using the multiscale Sato ridge filter and Gaussian scales $\sigma={1,2,3}$ pixels \cdh{which one?}\cite{SATO1998143}. \cdh{grammar?  I cannot parse this.}
% The resulting ridge-response image is rescaled to $[0,1]$, smoothed with a Gaussian filter with $\sigma=1$ pixel, and sharpened using an unsharp mask with a radius of 2 pixels and an amount of 2 \cdh{an amount?}. 
% A final intensity rescaling ensures a consistent range before segmentation. 
% Although the implementation supports alternative ridge filters and additional denoising or contrast-normalization operations, these options were not active for the results presented here.

\subsection{Automatic segmentation and morphological refinement.}
Segmentation is performed using triangle thresholding. 
For each enhanced image, an image specific threshold is calculated from its intensity histogram.
The selected threshold is illustrated in the final histogram of Figure~\ref{fig:mat:img_preproc}. 
Because the Sato response assigns larger values to detected ridge features, pixels with intensities greater than the threshold are classified as cracks. 
Triangle thresholding is well suited to these images because crack pixels constitute a relatively small population within the enhanced image, while the background forms the dominant histogram peak. 
% Under this class imbalance, variance based methods such as Otsu thresholding can produce unstable separation. 
\vr{Under this strong class imbalance, Otsu thresholding can be sensitive to small changes in the histogram and can shift the selected threshold toward the dominant background population.}
The triangle method instead determines the threshold geometrically from the shape of the histogram \cite{zack1977,otsu1979,sezgin2004}.

The resulting binary mask is refined by first removing connected components containing 10 pixels or fewer. 
This suppresses isolated responses caused by residual texture or noise before nearby crack segments can be joined. 
Morphological closing is then performed with a ($5\times5$) square structuring element to connect small discontinuities and produce a more continuous crack mask. 
The cleaned binary mask, shown as an output in Figure~\ref{fig:galaxy:results}, is used to calculate crack area fraction.

For topology based analysis, the cleaned mask is reduced to a one pixel wide skeleton and closed using the same ($5\times5$) structuring element. 
A structure tensor with $\sigma=2$ pixels is evaluated on the skeleton to identify locally directional features. 
Pixels are retained when their directional coherence satisfies
\begin{equation}
    \gamma =
    \frac{\lambda_1-\lambda_2}
    {\lambda_1+\lambda_2+10^{-10}} > 0.8,
\end{equation}
where ($\lambda_1$) and ($\lambda_2$) are the local structure tensor eigenvalues. 
This filtering suppresses weakly directional skeleton features while retaining elongated crack segments. 
The resulting skeleton can support measurements of crack length, connectivity, branching, and network topology and is also shown in Figure~\ref{fig:galaxy:results}. 
The same thresholding and morphological parameters were applied to all images retained for crack analysis without image specific tuning, and all retained images were processed successfully.

\subsection{Scale-aware quantitative outputs}
\begin{figure}[tb]
  \centering
  \includegraphics[width=0.9\textwidth]{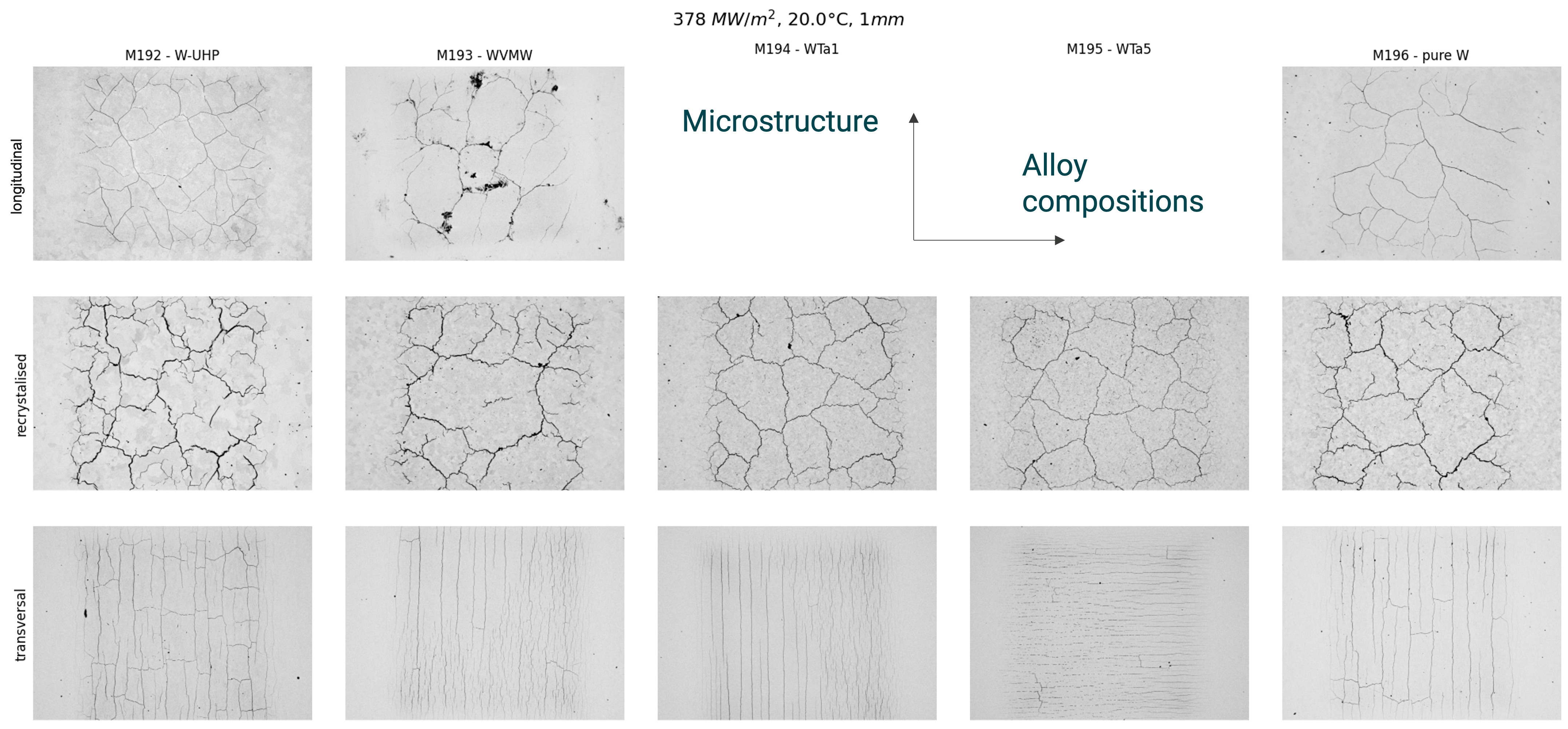}
  \rule{0.8\textwidth}{0.5pt}
  \\[1em]
  \includegraphics[width=0.9\textwidth, trim=0.4cm 0 0 2cm, clip]{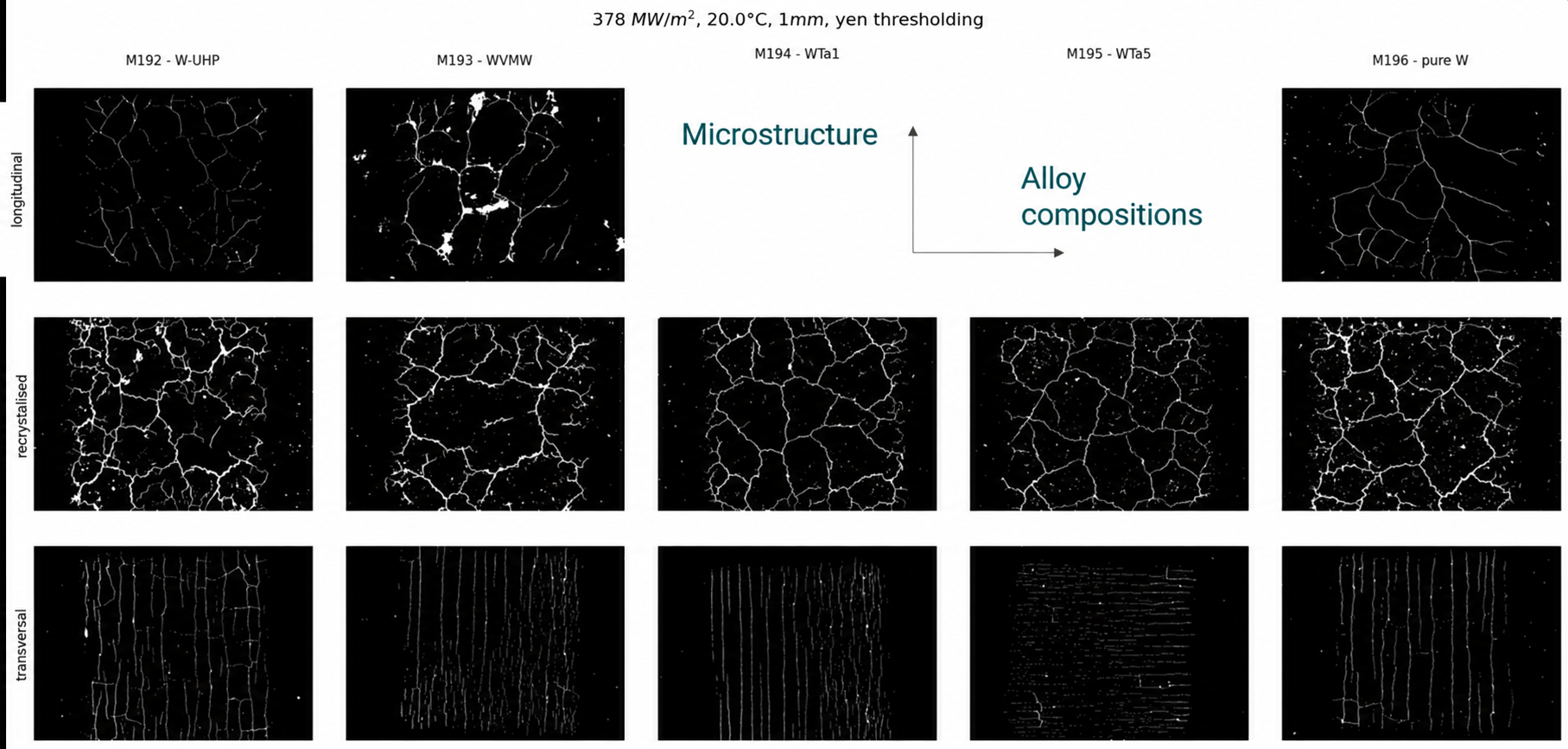}
  \caption{Top: representative experimental SEM characterization for a single heat-flux and base-temperature condition across alloys and microstructures. Bottom: corresponding automated crack extraction used by the workflow.}
  \label{fig:mat:sem}
\end{figure}

The primary scalar descriptor used in the current workflow is crack density,
$$
    \rho_{\mathrm{crack}} = \frac{N_{\mathrm{crack}}}{N_{\mathrm{valid}}},
$$
where $N_{\mathrm{crack}}$ is the number of pixels classified as crack and $N_{\mathrm{valid}}$ is the number of analyzed pixels after masking invalid or excluded regions. 
Crack density is therefore the fraction of the analyzed surface area identified as crack. 
When physical pixel size is available, the workflow also supports length density from skeletonized cracks and component-level statistics in physical units.
Examples of the resulting crack extraction are shown in Figure~\ref{fig:mat:sem}, demonstrating the ability of the pipeline to recover crack networks across different alloys and microstructures despite substantial variability in image appearance.

\begin{figure}[tb]
    \vspace{0pt}
    \centering
    \includegraphics[width=\textwidth]{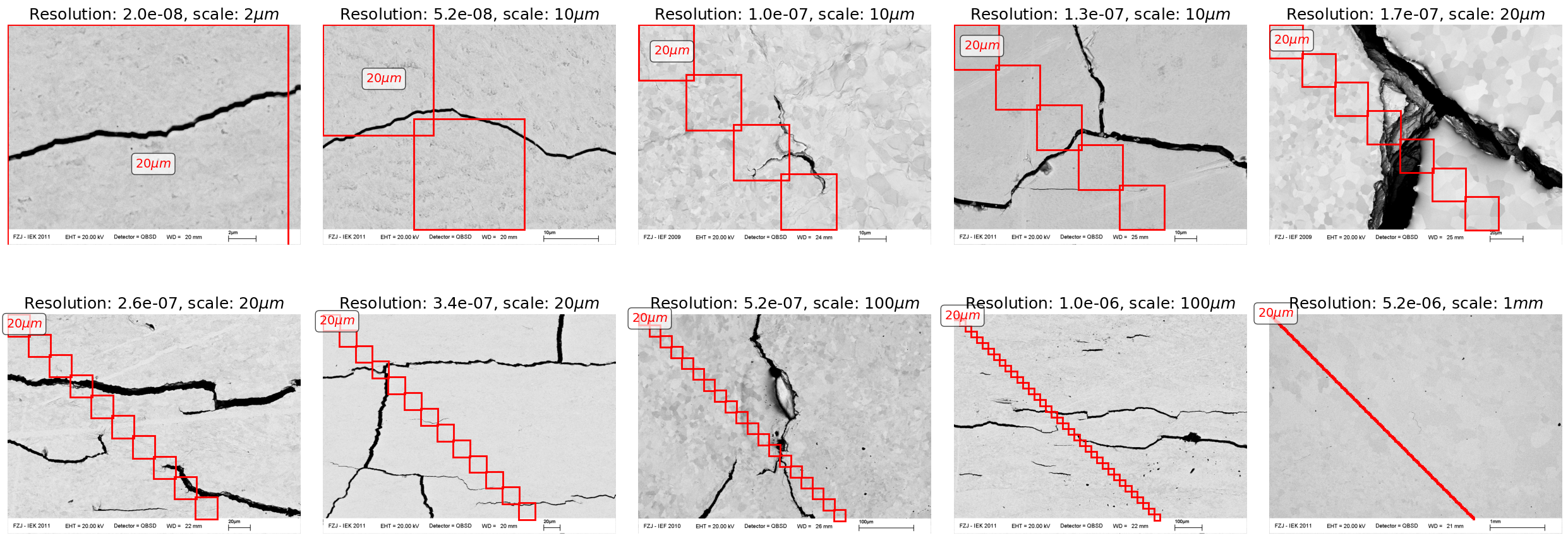}
    \rule{\textwidth}{0.5pt}
    \\[1em]
    \includegraphics[width=\textwidth]{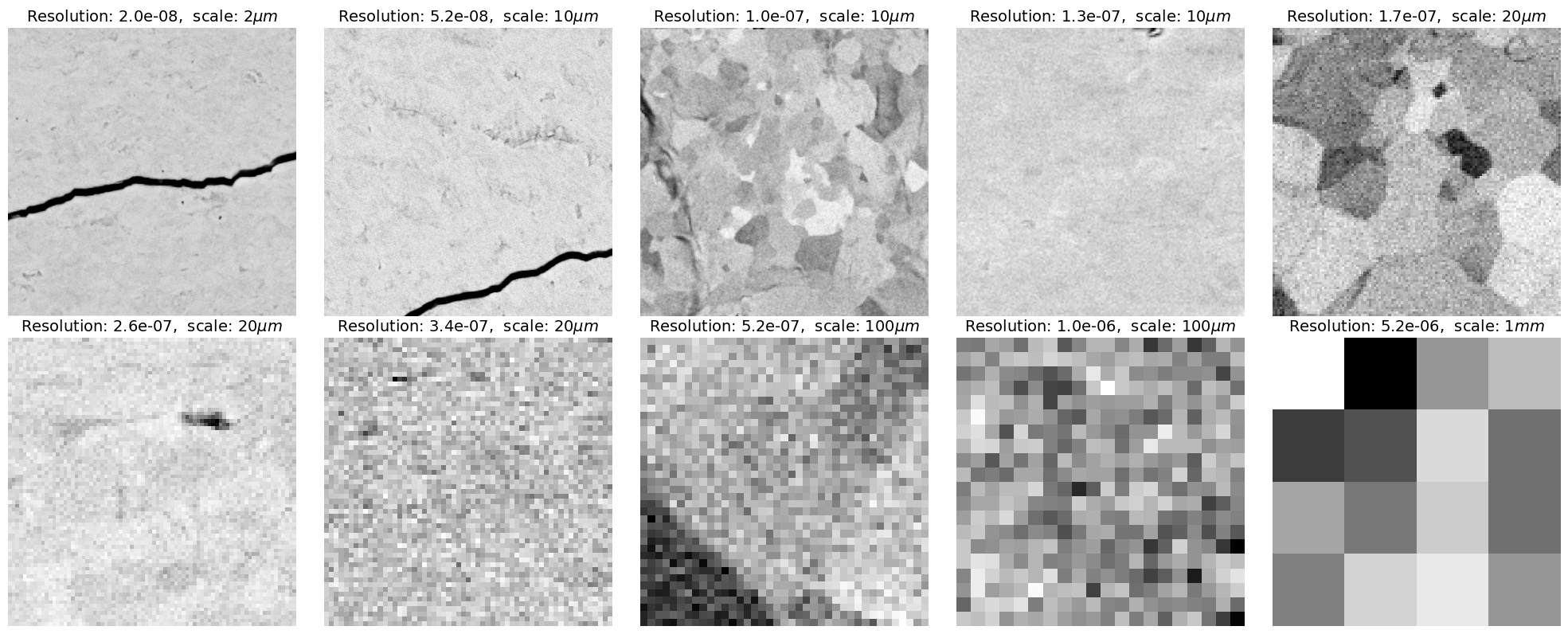}
    \caption{
    Top row: Representative SEM images acquired at different physical pixel resolutions, with overlaid patch boundaries corresponding to a fixed physical size of $20\mu\mathrm{m}$. Physical pixel resolution is defined as the physical length represented by one pixel. 
    Bottom row: Representative patches extracted from these images. Patches with coarse physical pixel resolution contain too few pixels across the fixed physical region to resolve crack and surface morphology reliably and are therefore excluded from morphology analysis.
    }
    \label{fig:mat:patches}
\end{figure}

Because crack area fraction depends on resolution and threshold-defined width, comparisons are made only among images with compatible physical resolution and processing profiles. 
The workflow retains image-scale metadata and flags images that are under-resolved for morphology analysis. 
For downstream representation learning, images are decomposed into patches defined by a fixed physical field of view rather than a fixed number of pixels. 
This preserves physically comparable crack morphology across different magnifications, see Figure~\ref{fig:mat:patches}.

\section{Workflow Demonstration and Quality Control}
\label{sec:workflow}
The Galaxy tool processes either individual images or collections of images. 
% For each image it produces the original and enhanced views, histogram and selected threshold, binary mask, overlay, optional skeleton, and a table of quantitative descriptors. 
For each image, it retains the original microscopy image, the enhanced image used for segmentation, the intensity histogram and automatically selected threshold, the binary crack mask, the detected cracks overlaid on the original image, the one pixel wide crack skeleton when topology based analysis is requested, and a table of quantitative descriptors.
These outputs are shown in Figure~\ref{fig:galaxy:results}.
They are retained rather than reduced immediately to a single descriptor so that experts can verify how the automated result was obtained and assess whether it is consistent with the surface features visible in the original image.
The tabular outputs support subsequent statistical analysis and model training.
% These products are intentionally retained rather than reduced immediately to a single number. 
% Visual products allow experts to distinguish true segmentation failures from scientifically unusual surfaces, while the tabular output supports statistical analysis and model training.

The pipeline recovers crack networks across multiple alloys and microstructures under heavily cracked conditions and also returns low crack area fractions for nominally undamaged images. 
Its principal advantage is consistency: the same processing logic and parameter values are applied across the dataset without image specific threshold tuning. 
% Galaxy records the exact parameter profile and software environment, making the result reproducible and allowing revised versions of the tool to be compared against prior histories \cite{GalaxyCommunity2024,Goecks2010}.
Galaxy records the input data, selected parameters, software environment, processing history, and generated outputs, making the analysis reproducible and allowing revised versions of the tool to be compared with earlier executions \cite{GalaxyCommunity2024,Goecks2010}.

Quality control is implemented at three levels.
% Input checks verify image readability and required metadata. 
Input checks verify that each image can be read and that the required image scale and experimental metadata are available.
% Algorithmic checks record the selected threshold, foreground fraction, connected-component statistics, and warnings for degenerate masks. 
Algorithmic checks record the selected threshold, the foreground fraction, defined as the fraction of pixels classified as cracks, and connected component statistics, including the number and sizes of separate detected regions.
The tool can also flag degenerate masks in which almost no pixels or almost all pixels are classified as cracks. 

% Visual checks provide side-by-side outputs for expert review. 
Visual checks present the original image, enhanced image, mask, overlay, and skeleton side by side for expert review. 
Images can therefore be retained with a quality flag rather than silently discarded, which is important for preserving the experimental record and for identifying new surface morphologies that fall outside the current algorithm's assumptions.

% \section{Preliminary Predictive and Simulation Modules}
\section{Interfaces to Downstream Modeling}
\label{sec:modeling}

% \subsection{Morphology-informed crack prediction}
% The automated masks and descriptors are used by a preliminary damage-prediction module. 
% To make the representation consistent across resolutions, SEM images are decomposed into patches defined by a fixed physical field of view. 
% Each patch is encoded with a pretrained DINOv2 vision transformer \cite{Oquab_2024_TMLR_DINOv2,Whitman_2024_CVPR,WhitmanLatypov_2025_ActaMaterialia}.
% Principal-component analysis is used for initial compression, followed by diffusion coordinates to obtain a low-dimensional representation in which distances reflect connectivity along the morphology manifold \cite{Jolliffe2002PCA,Coifman2006DiffusionMaps}.
% Figure~\ref{fig:mat:embeddings} illustrates this embedding pipeline.

\begin{figure}[tb]
    \vspace{0pt}
    \centering
    \includegraphics[width=\textwidth, trim=0 0 16.5cm 0, clip]{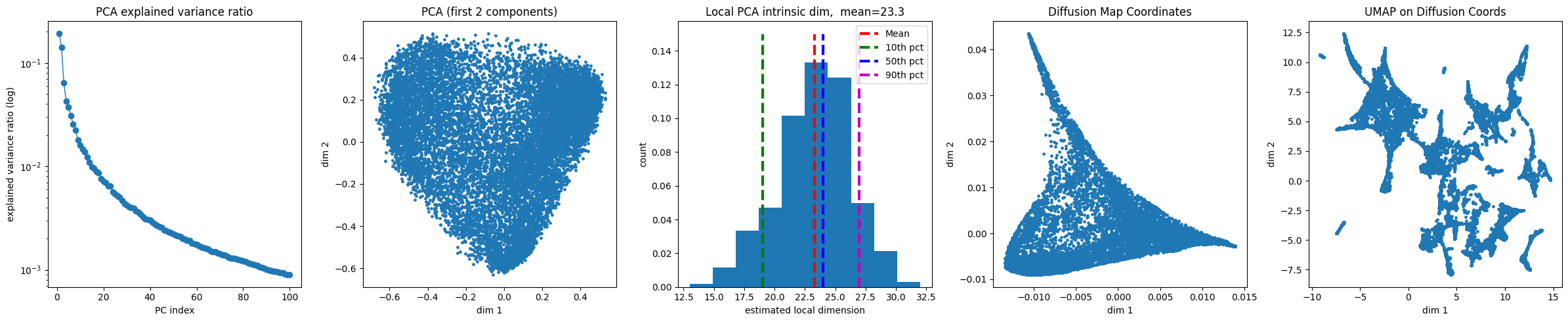}
    \caption{Embedding pipeline for DINOv2 patch features. 
    \vr{From left to right: PCA explained-variance decay used to assess linear compression; projection onto the first two principal components; distribution of local intrinsic-dimension estimates; and the first two diffusion-map coordinates illustrating nonlinear neighborhood structure.}}
    % From left to right: PCA variance decay, PCA projection, intrinsic dimension estimates, and diffusion map coordinates.}%, and UMAP visualization.}
    \label{fig:mat:embeddings}
\end{figure}

\subsection{Morphology informed prediction}

The standardized crack masks and quantitative descriptors produced by the workflow provide inputs for future damage prediction models. 
To support comparisons across image resolutions, SEM images are divided into patches with a fixed physical size. 
Each patch can then be represented using features from a pretrained vision model, such as DINOv2 \cite{Oquab_2024_TMLR_DINOv2}, followed by dimensionality reduction to obtain a compact description of surface morphology \cite{Whitman_2024_CVPR,WhitmanLatypov_2025_ActaMaterialia,Jolliffe2002PCA,Coifman2006DiffusionMaps}. 
Figure~\ref{fig:mat:embeddings} illustrates this representation pipeline.

Principal component analysis \vr{(PCA)} is first applied to remove redundant directions and provide a linear low dimensional baseline \cite{Jolliffe2002PCA}. 
Diffusion coordinates are then considered because the morphology features may vary along a nonlinear manifold. 
For example, gradual changes in crack density, branching, orientation, and surface texture can produce curved trajectories in feature space that may not be represented efficiently by a small number of linear principal components. 
Whereas PCA identifies directions of maximum global variance, diffusion coordinates emphasize local neighborhood relationships and connectivity among morphologically similar patches \cite{Coifman2006DiffusionMaps}. 
In this study, diffusion coordinates are included as an exploratory representation rather than as a demonstrated improvement over PCA.

The resulting morphology representations can be related to material composition, microstructure, and exposure conditions to support interpolation across the experimental test matrix. 
They may also provide a basis for uncertainty estimation and selection of informative future experiments or simulations. 
In the present work, this module is included only to demonstrate how the outputs of the crack identification workflow can be reused by downstream predictive tools. 
A systematic comparison of PCA, diffusion coordinates, and alternative embeddings, together with detailed model formulation, validation, and uncertainty quantification, is reserved for a follow on study.

\subsection{Interface to physics-based simulation}

\vr{The workflow provides an interface for incorporating results from physics-based fracture simulations into the same analysis framework used for experimental characterization.
Simulation outputs representing surface damage can be converted into image-like or descriptor-compatible products and processed through the same quantitative interface used for SEM data.
This establishes a common representation for experimental and simulated damage and enables simulation results to be compared with experimental observations within the Galaxy workflow.
The interface is designed to be independent of a particular simulation code, allowing different physics-based models to be incorporated without changing the upstream characterization workflow.

One candidate simulation capability is CabanaPD, an ORNL-developed peridynamic thermomechanical fracture code \cite{cabanapd}.
Peridynamics is a nonlocal reformulation of continuum mechanics that accommodates displacement discontinuities and crack evolution without constraining cracks to a finite-element mesh \cite{SILLING2005,GIANNAKEAS20183037,DANTUONO201774}.
CabanaPD is performance portable through Kokkos and Cabana and can execute on CPUs and GPUs, including leadership-class systems \cite{kokkos2022,slattery2022,cabanapd}.
In the present work, CabanaPD serves as an example of a physics-based simulation capability that can be connected through the workflow interface; its application to tungsten thermal-shock conditions, including material parameterization, model developments, calibration, and validation, will be evaluated separately.
The present manuscript therefore focuses on establishing the computational interface between experimental characterization and physics-based simulation rather than on the formulation or validation of a specific fracture model.
}

\section{Discussion}
\label{sec:discussion}

\subsection{The workflow as the scientific contribution}
Automated segmentation algorithms are often reported independently of the data and computing processes required for practical use.
For fusion-materials research, however, utility depends on the full chain from acquisition to decision. 
The same code can produce non-reproducible results if image calibration is lost, software versions differ, parameters are modified without record, or outputs are transferred manually.
Galaxy makes the workflow itself a persistent scientific object: tools have explicit interfaces, executions are recorded in histories, data types constrain valid connections, and workflows can be shared and re-run \cite{NievedelaHidalgo2025ROCrates,GalaxyCommunity2024,Goecks2010}.

This architecture also separates component maturity. The crack-identification tool has been prototyped and deployed in Galaxy and is the most developed component described here.
The full material-assessment workflow, including prediction, uncertainty-driven sampling, and automated simulation dispatch, remains an integration target. 
Stating that distinction explicitly avoids conflating a demonstrated image-analysis capability with a fully validated digital twin.

\subsection{Interpretation and limitations}
Crack area fraction is useful for comparing surface damage, but it is not a complete description of fracture. It depends on image resolution, segmentation width, detector contrast, and the selected physical field of view. 
Skeleton-based length density, component size, orientation, connectivity, and branching can provide complementary information. 
These descriptors are supported by the workflow data model and can be added without changing the upstream segmentation interface.

The present image processing pipeline is optimized for elongated crack features.
Strong grain boundaries, polishing lines, or elongated pores can be misidentified when they closely resemble cracks, whereas wide spalled regions or diffuse surface modification may not produce a ridge-like response~\cite{Zhao2021SEMCrack,MohanPoobal2018}.
These limitations motivate future multi-class characterization in which cracking, roughening, melting, erosion, and redeposition are treated as distinct but related damage modes.

The predictive module is constrained by both data coverage and incomplete materials metadata. 
Current inputs include exposure conditions and coarse material identifiers, but do not yet fully encode grain-size distributions, crystallographic texture, \vr{temperature-dependent material properties, or anisotropy.} %do not yet fully encode grain-size distributions, crystallographic texture, temperature-dependent conductivity, elastic modulus, yield strength, or anisotropy. 
A learned metric can partially compensate for missing structure but cannot recover unobserved physics. The simulation module has complementary limitations because it currently represents continuum-level anisotropy rather than explicit microstructure.

\begin{figure}[tb]
  \centering
  \includegraphics[width=\textwidth]{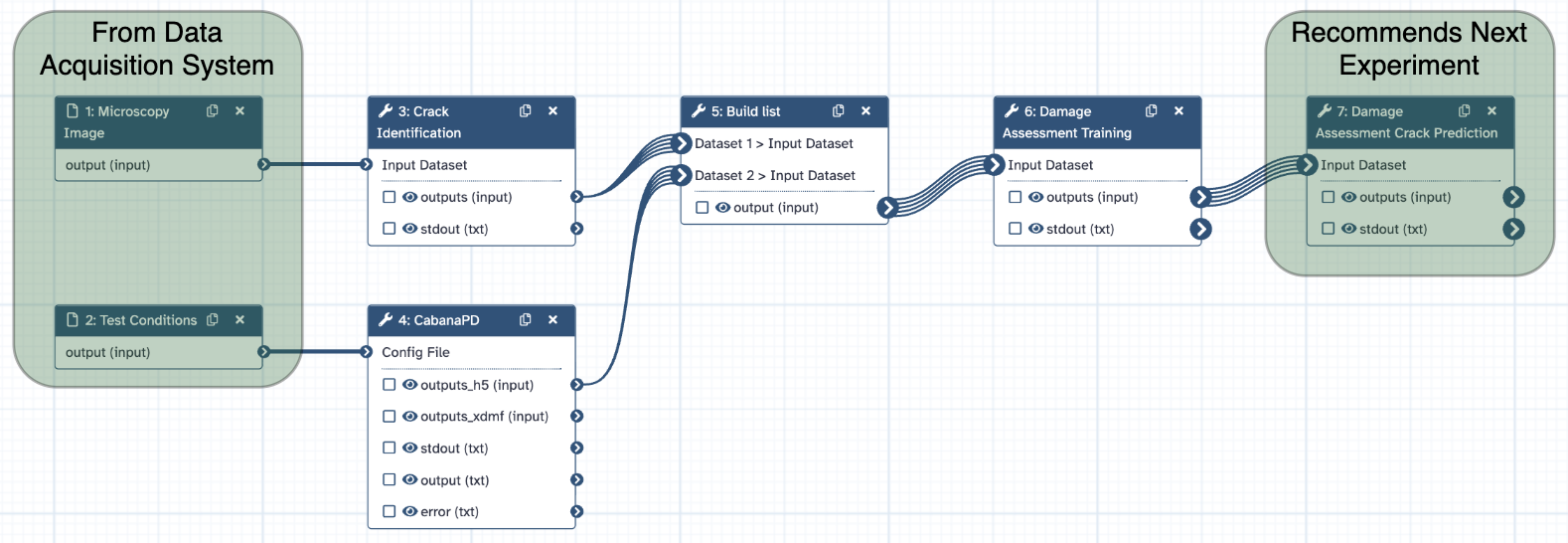}
  \caption{Example from the Galaxy workflow builder showing the end goal - a fully reproducible experiment steering workflow.}
  \label{fig:galaxy:workflow}
\end{figure}

\subsection{Path toward closed-loop material assessment}
The long-term architecture is the closed loop shown in Figure~\ref{fig:galaxy:workflow}.
Experimental images are ingested and processed automatically; the surrogate estimates damage and uncertainty over material and exposure space; an acquisition strategy identifies the most informative new experiment or simulation; Galaxy dispatches the selected computation or ingests new experimental data; and the model is updated~\cite{Lookman2019ActiveLearning,RasmussenWilliams2006GPML}.
The immediate value of the present work is to establish the reproducible characterization layer and the interfaces required for that loop.

The same workflow concepts extend beyond the electron-beam dataset. 
In the MPEX Surface Analysis Station, SEM, focused-ion-beam sections, spectroscopy, and other modalities can be combined through typed Galaxy tools and provenance-tracked histories~\cite{BiewerRapp2020SAS,GalaxyCommunity2024}. The crack-identification module therefore serves as the first operational component of a broader MPEX material-assessment workflow.

\section{Conclusions}
We developed a reproducible Galaxy workflow for automated crack identification and quantitative post-exposure assessment of fusion materials.
The workflow combines scale-aware data ingestion, automated image enhancement and segmentation, quality control, descriptor extraction, provenance capture, and \vr{modular interfaces to data-driven and physics-based models}. %modular interfaces to predictive and physics-based models. 
% Its core crack-identification method uses Sato ridge enhancement, Gaussian histogram conditioning, unsharp masking, triangle thresholding, and morphological refinement to process heterogeneous SEM images without image-specific parameter tuning.

Applied to 418 images from 114 electron-beam experiments on five tungsten grades and three microstructural states, the workflow produces crack masks, overlays, skeletons, and crack-density descriptors across highly cracked, weakly damaged, and nominally undamaged surfaces. 
Galaxy deployment makes the analysis shareable, re-executable, and portable across computing resources while preserving every input, parameter, software version, and output in an execution history.

% Preliminary morphology-informed regression and CabanaPD modules demonstrate how standardized characterization outputs can support crack prediction and simulation-guided exploration of sparse material and exposure spaces. 
% These components are deliberately presented at high level and will be addressed in dedicated follow-on studies.
% The central result is an extensible workflow that converts microscopy images into provenance-tracked quantitative data and establishes the computational foundation for closed-loop, AI-assisted fusion-materials assessment.

\vr{
Preliminary morphology-informed prediction and interfaces to physics-based simulation illustrate how standardized characterization outputs can support crack prediction and comparison with simulated damage across sparsely sampled material and exposure spaces. These downstream capabilities are presented at a high level, while detailed predictive-model development and the evaluation, calibration, and validation of specific simulation approaches are reserved for dedicated follow-on studies. The central result is an extensible workflow that converts microscopy images into provenance-tracked quantitative data and establishes a computational foundation for closed-loop, AI-assisted fusion-materials assessment.
}

\section*{Acknowledgments}

The authors thank Marius Wirtz and Jan W. Coenen for access to JUDITH electron-beam data and Sam Reeve and Pablo Seleson for scientific discussions.

\bibliographystyle{style/ans_js}
\bibliography{bibliography}

\end{document}